%% file: template.tex
\documentclass[review]{fcs}

\usepackage{xspace}

\newcommand{\eg}{e.g.\xspace}

\usepackage{times}
\usepackage{latexsym}
\usepackage{amsmath}

\usepackage{tabularx}  
\usepackage{booktabs}

\usepackage[T1]{fontenc}

\usepackage[utf8]{inputenc}
\usepackage{microtype}

\usepackage{inconsolata}

\usepackage{amssymb}
\usepackage{graphicx}
\usepackage{multirow}
\usepackage{algorithm}
\usepackage{algorithmic}
\usepackage{enumitem}
\usepackage{xcolor}
\usepackage{subfigure}
\usepackage{diagbox} 
\usepackage{amsmath}
\usepackage{amsfonts}

\usepackage[edges]{forest}
\usepackage{tabularx}
\usepackage{makecell}

\usepackage{longtable}
\usepackage{booktabs}
\usepackage{ragged2e}
\usepackage{array}

\newcolumntype{J}{>{\RaggedRight\arraybackslash}X}           
\newcolumntype{K}{>{\RaggedRight\arraybackslash}X}           
\newcolumntype{N}{>{\centering\arraybackslash}m{1.8cm}}      
\newcolumntype{U}{>{\RaggedRight\arraybackslash}m{0.25\textwidth}} 

\title{Fast, Slow, and Tool-augmented Thinking for LLMs: A Review}
\author[1]{Xinda Jia}
\author[3]{Jinpeng Li}
\author[4]{Zezhong Wang}
\author[5]{Jingjing Li}
\author[6]{Xingshan Zeng}
\author[]{Yasheng Wang}
\author[2]{Weinan Zhang}
\author[2]{Yong Yu}
\author*[2]{Weiwen Liu}  

\address[1]{School of Mechanical Engineering, Shanghai Jiao Tong University, Shanghai 200240, China}
\address[2]{School of Computer Science, Shanghai Jiao Tong University, Shanghai 200240, China}
\address[3]{Huawei Technologies Co., Ltd., Beijing 100084, China}
\address[4]{Department of Systems Engineering and Engineering Management, The Chinese University of Hong Kong, Hong Kong 999077, China}
\address[5]{Department of Computer Science and Engineering, The Chinese University of Hong Kong, Hong Kong 999077, China}
\address[6]{Huawei Hong Kong Research Center, Hong Kong 999077, China}

\fcssetup{
  received       = {month dd, yyyy},
  accepted       = {month dd, yyyy},
  corr-email     = {wwliu@sjtu.edu.cn},
}

\begin{abstract}
Large Language Models (LLMs) have demonstrated significant progress in reasoning across diverse domains. 
However, effective reasoning in real-world tasks requires adapting both the computational depth and the knowledge source to the demands of the problem, ranging from fast, intuitive responses to deliberate, step-by-step reasoning and from purely internal reasoning to externally tool-augmented processes.
Drawing inspiration from cognitive psychology, we propose a novel taxonomy of LLM reasoning strategies along two orthogonal knowledge boundaries: (1) a fast/slow boundary separating intuitive from deliberative processes, and (2) an internal/external boundary distinguishing reasoning grounded in the model’s parameters from reasoning augmented by external tools. 
We systematically survey recent work on adaptive reasoning in LLMs and categorize methods based on key decision factors. 
We conclude by highlighting open challenges and future directions toward more adaptive, efficient, and reliable LLMs.
\end{abstract} 

\begin{document}

\input{intro}

\input{def}
\input{cog}

\input{imex}

\input{FCS-251673-fig3}
\input{knowledgeboundary}

\input{future}

\begin{acknowledgement}
The work is supported by National Natural Science Foundation of China (62502310,62322603).
\end{acknowledgement}

\bibliographystyle{fcs}
\bibliography{custom}

\end{document}

%% file: intro.tex
\begin{figure*}[t]
    \centering
    {\captionsetup{skip=2pt, belowskip=0pt}
    \includegraphics[width=\textwidth]{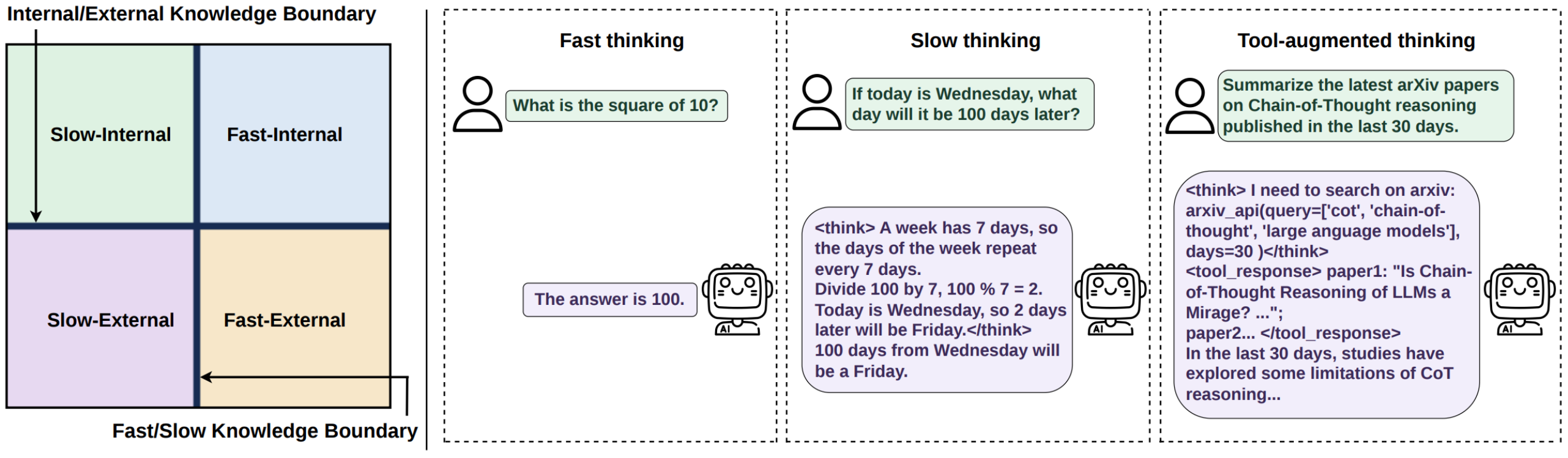}
    \caption{Comparison of fast, slow, and tool‑augmented thinking. The left part shows the two knowledge boundaries that define the $2\times 2$ structure of reasoning strategies: the fast/slow knowledge boundary and the internal/external knowledge boundary. The right part showcases representative tasks for each reasoning strategy.}
    \label{fig:overall}}
\end{figure*}

\section{Introduction}

\begin{quote}
\textit{The mind is not a single, undifferentiated entity, but a collection of specialized mechanisms shaped by evolution to solve different categories of problems.}
\hfill --- Steven Pinker, \textit{How the Mind Works} (1997)
\end{quote}

Large Language Models (LLMs) have been advancing rapidly, steadily pushing the boundaries toward Artificial General Intelligence (AGI)~\cite{anthropic2025claude,guo2025deepseek}.
Significant breakthroughs have been made in LLM reasoning capabilities, including mathematics, coding, and logical reasoning, as seen in models like OpenAI o3~\cite{openai2025o3}, DeepSeek-R1~\cite{guo2025deepseek}, and Grok 4~\cite{xai2025grok4}. 
Reasoning---the ability to draw inferences, make decisions, and solve problems based on available information---is central to human intelligence~\cite{li2502system}. 
Recent research has identified several distinct reasoning strategies in LLMs, loosely aligning with theories of human cognition~\cite{kahneman2011thinking}. 

In particular, these strategies are often discussed in terms of fast, slow, and tool-augmented thinking.
From a structural perspective, these reasoning strategies can be characterized with two orthogonal dimensions rather than as three mutually exclusive modes. 
The fast/slow knowledge boundary reflects differences in computation depth, distinguishing intuitive, rapid responses from deliberate, step-by-step reasoning. The internal/external knowledge boundary captures the differences in the knowledge sources, distinguishing reasoning grounded solely in the internal parameters of a model from reasoning augmented by external tools or information sources. In this view, tool-augmented thinking is not a standalone alternative to fast or slow thinking, but an external knowledge dimension that can intersect with both. 
\noindent\textbullet\ \textbf{Fast thinking:}\par
\noindent
Analogous to human System 1 cognition, fast thinking involves intuitive responses generated directly from the model without explicit intermediate steps~\cite{pan2024dynathink}. 
Fast thinking offers low latency and high throughput, making it particularly suitable for simple or routine tasks such as casual conversation, sentiment classification, surface-level summarization, and factual questions with high model confidence. However, it is prone to errors when the input is ambiguous, unfamiliar, or requires multi-step reasoning.

\noindent\textbullet\ \textbf{Slow thinking:}\par
\noindent
Similar to human System~2 cognition, slow thinking engages in step-by-step, deliberative reasoning.
Techniques such as chain-of-thought (CoT) prompting~\cite{wei2022chain} and its extensions like the tree-of-thoughts (ToT)~\cite{yao2023tree}, as well as other techniques including self-reflection~\cite{ji2023towards} and intermediate verification~\cite{liang2024improving}, enable models to decompose problems, explore alternative reasoning paths, and validate intermediate results.
This mode is particularly effective for complex tasks such as mathematical problem solving, logical deduction, planning, and scientific explanation, where intermediate reasoning steps are crucial for correctness, although at the cost of increased latency and computation.



    
\noindent\textbullet\ \textbf{Tool-augmented thinking:}\par
\noindent
Tool-augmented thinking extends beyond the model’s innate capabilities by incorporating external tools (\eg, calculators, code interpreters, search engines or knowledge databases). 
This mode is especially useful when tasks require precise numerical computation, access to up-to-date or domain-specific knowledge, or interaction with external environments, such as web search, code execution, and retrieval-augmented generation (RAG). Tools can support both fast thinking, by providing quick access to relevant information, and slow thinking, by adding depth through detailed computations, intermediate verification, and exploration of alternative reasoning paths. This mirrors humans' reliance on external aids to compensate for memory limitations and knowledge gaps, enabling both faster responses and more thorough reasoning, depending on the task at hand.

However, the complexity and diversity of real-world applications demand that LLMs flexibly adapt their reasoning strategy to each task’s accuracy and latency constraints.
Hence, selecting and shifting between the appropriate reasoning strategies of LLMs becomes an active and growing area of research~\cite{zhao2025let,kojima2022large}. 
Different tasks call for different reasoning modes: fast, intuitive responses may suffice for casual conversation or surface-level summarization, while complex problem solving, multi-step logical deduction, or tasks requiring factual precision often necessitate slower, more deliberate reasoning or external tool integration. 
A mismatch between task demands and the model’s reasoning strategy can lead to inefficiencies or errors~\cite{sun2025detection}.

For example, underthinking can lead to superficial or incorrect responses when a deeper analysis is needed, such as mathematical problem solving, legal argumentation, or scientific explanation~\cite{wei2022chain}.
On the other hand, overthinking can also be detrimental~\cite{schuster2022confident}. 
Applying complex reasoning strategies to simple tasks can introduce unnecessary latency and compute, and intermediate traces may contain hallucinated or spurious steps that mislead the final answer~\cite{sun2025detection}.
Moreover, language models that rely solely on internal knowledge are prone to hallucination, especially on factual or long‑tail queries~\cite{roberts2020much}. Yet blindly incorporating retrieved knowledge can destabilize model behavior and propagate errors~\cite{moskvoretskii2025adaptive}.

\textbf{\textit{Developing a deeper understanding of how LLMs reason, and how these reasoning processes can be guided or improved}}, is essential for both advancing model capabilities and enhancing their reliability in real-world applications.
To address this need, we propose a taxonomy structured along two knowledge boundaries, as shown in Figure~\ref{fig:overall}. The fast/slow knowledge boundary reflects differences in reasoning depth, distinguishing intuitive, rapid responses from deliberate, step-by-step reasoning. The internal/external knowledge boundary captures differences in knowledge sources, distinguishing reasoning grounded solely in a model’s internal parameters from reasoning augmented by external tools or information sources.


Building on these two knowledge boundaries, we further survey and categorize existing research on adaptive reasoning strategy selection. To the best of our knowledge, this is the first comprehensive review that systematically examines the adaptive selection among all three major reasoning strategies---fast thinking, slow thinking, and tool-augmented thinking. Our main contributions are summarized as follows:
\begin{itemize}
    \item We present the first comprehensive taxonomy of reasoning strategies in LLMs, organized along two knowledge boundaries: the fast/slow knowledge boundary and the internal/external knowledge boundary.
    \item We conduct a systematic analysis of existing research on adaptive reasoning strategy selection, categorizing methods based on key decision factors such as model confidence, task complexity, and utility gain.
    \item We identify and discuss open challenges and promising future directions for improving reasoning strategy selection in LLMs.
\end{itemize}

%% file: def.tex
\section{Formulation}
\label{sec:def}
Inspired by the decision-making principles widely adopted in adaptive reasoning systems, we view reasoning strategy selection in LLMs as a process of evaluating multiple decision signals and crossing the boundaries. Accordingly, we introduce a unified framework that abstracts existing methods into decision factor extraction and score-based strategy selection, without assuming a specific implementation.

\subsection{Decision Factor Extraction}
Given an input query $x$ and model parameters $\theta$, a reasoning strategy selection framework for LLMs first extracts some decision factors. 

These are represented as:
\begin{align}
    \mathbf{g}^{(d)}(x,\theta,c)=[f_1^d, f_2^d,\ldots, f_{n_d}^d]\label{eq:1}, \\ 
    \mathbf{g}^{(s)}(x,\theta,c)=[f_1^s, f_2^s,\ldots, f_{n_s}^s],\label{eq:2}
\end{align}
Here, $\mathbf{g}^{(d)}$ and $\mathbf{g}^{(s)}$ denote the lists of features relevant to reasoning depth and knowledge source selection, respectively. 
The optional control signal $c$ captures explicit guidance or constraints—such as user prompts, predefined rules, compute budgets, or external routing directives. In some scenarios, $c$ may be omitted or implicitly encoded within the model architecture or prompt.

The extracted decision factors can capture a range of behavioral signals from the model. These may include internal signals such as model confidence or uncertainty, as well as external contextual cues like task type or user intent. For example, the model's confidence distribution~\cite{schuster2022confident} measured via output logits serves as a proxy for uncertainty and can guide reasoning depth. Self-Route~\cite{he2025self} incorporates query complexity as a model-conditioned decision factor. By training on datasets with a continuous gradient of difficulty, its routing mechanism learns to adaptively determine whether a given query can be addressed with shallow reasoning or requires deeper processing.

\subsection{Strategy Selection}
Based on the extracted decision factors, the system applies a selection procedure to compute decision scores at one or more decision units:
\begin{align}
\boldsymbol{\Phi}_{d}(x, \theta, c) &= \mu_d\left( \mathbf{g}^{(d)}(x, \theta, c)\right)\,,\label{eq:phi_d}\\
\boldsymbol{\Phi}_{s}(x, \theta, c) &= \mu_s\left( \mathbf{g}^{(s)}(x, \theta, c) \right)\,,\label{eq:phi_s}
\end{align}
where $\boldsymbol{\Phi}_{d}$ and $\boldsymbol{\Phi}_{s}$ denote vectors of decision scores. They may contain one or multiple entries, depending on whether the strategy selection procedure $\mu_d$ or $\mu_s$ is performed once globally~\cite{lou2025adacot,liang2025thinkswitcher} or dynamically during generation, such as at the token, sentence or step level~\cite{jiang2023active,yan2025mur}.
Specifically, $\mathbf{g}^{(d)}$ or $\mathbf{g}^{(s)}$ may contain logit-derived confidence or uncertainty signals, and the corresponding selection procedure $\mu_d$ or $\mu_s$ can be applied in a sentence-wise manner. In this case, each sentence-level decision unit produces one decision score, and these scores together form the vector $\boldsymbol{\Phi}_d$ or $\boldsymbol{\Phi}_s$.
For example, in FLARE~\cite{jiang2023active}, $\mathbf{g}^{(s)}$ is instantiated by token probabilities derived from the model logits, and $\mu_s$ operates as a sentence-wise selection procedure. For each temporarily generated sentence, it records the generation probability of each token and computes a sentence level selection score based on the lowest token probability. The collection of these sentence-level scores forms the vector $\boldsymbol{\Phi}_s$.


Importantly, each entry in these vectors corresponds to a scalar decision criterion, denoted by $\Phi_{d,t}$ or $\Phi_{s,t}$. In some methods, this score is explicitly exposed and can take various forms, including linear aggregation, minimum over a set of values, confidence-based scores, or other rule-based decision criteria~\cite{liang2025thinkswitcher}. In other methods, especially end-to-end learned methods, it may not be directly reported but can be interpreted as an implicit preference induced by the learned policy over alternative reasoning actions~\cite{zhang2025adaptthink}.

Such selection procedures can be manually designed, specified heuristically, or learned via supervised training or reinforcement learning~\cite{tu2025learning}.
For example, the CGR framework~\cite{nogueira2025certainty} instantiates an adaptive reasoning depth selection framework in a step wise manner. It uses token probabilities derived from the model logits as confidence decision factors.
The selection procedure evaluates these factors at each predefined reasoning step by probing the current reasoning trace and produces a scalar uncertainty score, derived from the minimum token probability among the selected answer tokens, which is then compared against a predefined threshold to determine whether deeper reasoning should be triggered.

Specifically, $\boldsymbol{\Phi}_d$ and $\boldsymbol{\Phi}_s$ are vectors of selection scores used as routing variables, representing the expected benefit of allocating additional reasoning or external resources. The corresponding threshold vectors are denoted by $\boldsymbol{\delta}_d$ and $\boldsymbol{\delta}_s$, and the resulting selection outcomes are denoted by the vectors $\boldsymbol{k}_d$ and $\boldsymbol{k}_s$. For each decision unit $t$, when the corresponding score entry exceeds its associated threshold entry $\delta_{d,t}$ for reasoning depth or $\delta_{s,t}$ for knowledge source, the corresponding knowledge boundary is crossed, yielding the following decision functions:
\begin{equation}\label{eq:5}
k_{d,t}(x,\theta, c) = 
\begin{cases}
\textit{fast thinking}, \text{if } \Phi_{d,t} < \delta_{d,t} \\
\textit{slow thinking}, \text{if } \Phi_{d,t} \geq \delta_{d,t}
\end{cases}\\
\end{equation}
\begin{equation}\label{eq:6}
k_{s,t}(x, \theta, c) = 
\begin{cases}
\textit{internal knowledge},\text{if } \Phi_{s,t} < \delta_{s,t} \\
\textit{external knowledge},\text{if } \Phi_{s,t} \geq \delta_{s,t}
\end{cases}
\end{equation}
where the threshold vectors $\boldsymbol{\delta}_d$ and $\boldsymbol{\delta}_s$ are typically model-dependent.
The implementation of this threshold mechanism varies significantly across the surveyed literature. In some approaches, each entry of the corresponding threshold vector is instantiated as an explicit scalar value that is either manually specified~\cite{yang2025dynamic, yao2024seakr} or empirically calibrated on a validation set~\cite{li2025adaptive}. In these cases, overall system performance is highly sensitive to the chosen threshold, as it directly controls the frequency of strategy switching and governs the critical trade-off between computational efficiency and reasoning accuracy. Conversely, other methods realize this switching criterion implicitly through training, such as through post-training alignment or learned gating networks~\cite{lou2025adacot, wang2024adaptive}. Within our framework, these learned routing mechanisms function as implicit decision boundaries that serve the same role as the threshold vectors $\boldsymbol{\delta}_d$ and $\boldsymbol{\delta}_s$.

It is worth noting that most existing studies that decide whether to use a particular reasoning mode usually treat the fast/slow and internal/external dimensions separately. However, these two dimensions may still interact in practical reasoning systems. For example, when a model decides to use external knowledge or tools, this result can further serve as a condition for deciding how much reasoning should be performed around the tool call. Such explicit coupling, however, remains relatively underexplored in existing work. Therefore, this interaction can be regarded as a potential direction for future extensions of our framework. Specifically, Eq. (5-6) can be extended by taking the internal/external selection result as an additional input to the fast/slow decision, thereby allowing the two dimensions to be modeled in a more coupled manner.
Finally, the model generates the response conditioned on the selected strategy:
\begin{equation}\label{eq:7}
    y = f_\theta(x, \boldsymbol{k}_d(x, \theta, c), \boldsymbol{k}_s(x,\theta, c))\,,
\end{equation}
where the vectors $\boldsymbol{k}_d$ and $\boldsymbol{k}_s$ denote the reasoning depth and knowledge source selection results across the decision units, respectively, and each can reduce to a single selection when only one decision unit is used for the corresponding dimension.

We next instantiate this framework with three representative methods, covering both reasoning depth selection and knowledge source selection. For example, MUR~\cite{yan2025mur} provides a concrete instantiation of adaptive selection between fast thinking and slow thinking. In MUR, fast thinking means directly proceeding with the current reasoning step, while slow thinking is implemented through test-time scaling (TTS). 
It first extracts step-level uncertainty $m_t$, derived from token probabilities, as the decision factor $\mathbf{g}^{(d)}$ in Eq.~\eqref{eq:1}. The selection procedure $\mu_d$ in Eq.~\eqref{eq:phi_d} aggregates the previous momentum uncertainty $M_{t-1}$ and the current uncertainty $m_t$ into 
$M_t=\alpha M_{t-1}+(1-\alpha)m_t$, 
which serves as an entry of decision score vector $\boldsymbol{\Phi}_d$ in Eq.~\eqref{eq:phi_d}. MUR then compares this score with a manually designed threshold each turn, such as $\exp(M_{t-1})/\gamma$, to determine the reasoning depth decision in Eq.~\eqref{eq:5}: if the uncertainty exceeds the threshold, the model switches to slow thinking by applying additional TTS computation; otherwise, it remains in fast thinking. Finally, the outputs of all reasoning steps are concatenated into the final response, corresponding to Eq.~\eqref{eq:7}.
As an example of knowledge source selection, FLARE~\cite{jiang2023active} instantiates adaptive selection between internal and external knowledge through confidence based active retrieval. Specifically, FLARE first generates a temporary next sentence $\hat{s}_t$ without retrieval and extracts token-level confidence scores as the decision factors $\mathbf{g}^{(s)}$ in Eq.~\eqref{eq:2}. The selection procedure $\mu_s$ in Eq.~\eqref{eq:phi_s} is instantiated as a sentence-wise confidence based algorithm: for each temporarily generated sentence, it computes the minimum token probability $\min_{w_i \in \hat{s}_t} p(w_i)$. In the original formulation, retrieval is triggered if any token probability falls below a manually specified confidence threshold $\theta$. Equivalently, by defining an uncertainty-based decision score $\Phi_{s,t} = 1 - \min_{w_i \in \hat{s}_t} p(w_i)$ and setting $\delta_{s,t} = 1-\theta$, the decision can be expressed in the form of Eq.~\eqref{eq:6}: if $\Phi_{s,t} \leq \delta_{s,t}$, the model accepts the internally generated sentence without retrieval; otherwise, it invokes external retrieval and regenerates the sentence conditioned on the retrieved documents. Through this sentence-level selection process, the final response is generated as a trajectory of internal and external knowledge-use decisions, corresponding to Eq.~\eqref{eq:7}.

Furthermore, AdaptThink~\cite{zhang2025adaptthink} illustrates a learned implicit form of reasoning depth selection, which corresponds to the scalar case of our vectorized formulation. The decision factors $\mathbf{g}_d$ in Eq.~\eqref{eq:1} correspond to signals indicating the necessity of explicit reasoning, such as problem difficulty and the relative utility of \textit{Thinking} over \textit{NoThinking}. The selection procedure $\mu_d$ in Eq.~\eqref{eq:phi_d} is instantiated by an RL-trained policy that learns to choose between the two modes. In its PPO-style objective, AdaptThink uses the advantage term $A(x,y)=\mathbb{I}(y_1=\texttt{</think>})\cdot\delta+R(x,y)-\bar{R}_{\mathrm{ref}}(x)$, where $\delta$ serves as a threshold-like control parameter encouraging \textit{NoThinking} when performance is not degraded. After training, the learned policy induces an implicit scalar preference score $\Phi_d$ in Eq.~(3), and determines the reasoning-depth result $k_d$ in Eq.~\eqref{eq:5}: generating \texttt{</think>} at the beginning corresponds to fast thinking, whereas continuing explicit reasoning corresponds to slow thinking. Therefore, AdaptThink realizes Eq.~\eqref{eq:5} through an implicit decision boundary learned from its RL objective, and the resulting strategy selection conditions the final response generation in Eq.~\eqref{eq:7}.

This unified two-stage framework, comprising decision factor extraction and strategy selection, offers a principled and modular approach to adaptive reasoning in LLMs. It separates the \textit{what to consider} (captured by decision factors) from the \textit{how to decide} (captured by selection functions), enabling more interpretable and controllable model behavior.

Prior work has explored different aspects of this framework. Some focus on improving decision factor extraction, for example by designing richer behavioral signals or learning better input representations~\cite{pan2024dynathink,jiang2023active}. Others emphasize the design of strategy selection mechanisms~\cite{lou2025adacot,zhang2025adaptthink}, leveraging methods such as rule-based heuristics, supervised learning, or reinforcement learning to map decision factors to optimal reasoning modes.
Our two-stage framework successfully accommodates a broad spectrum of adaptive reasoning methodologies, specifically capturing approaches that emphasize decision factor extraction~\cite{zhu2025uncertainty, zhang2025reasoning, zubkova2025sugar}, strategy selection mechanisms~\cite{liang2025thinkswitcher,zhao2025let,he2025self}, or a combination of both~\cite{zhang2025adaptthink, pan2024dynathink, yan2025mur, nogueira2025certainty}.
Conversely, a primary boundary condition of our formulation is that it does not encompass \textit{reasoning strategy fusion} paradigms. Methods such as SMaRT~\cite{verma2025smart}, which concurrently employ multiple reasoning strategies and combine their outputs to reach a final conclusion, operate on an ensemble principle rather than a routing mechanism. Consequently, our framework should be interpreted as a unifying abstraction designed specifically for selection-based adaptive reasoning, rather than fusion- or aggregation-based approaches.

%% file: cog.tex
\section{Perspective of Cognitive Psychology}

Understanding the reasoning strategies of LLMs through the lens of cognitive psychology offers a powerful basis for both interpreting their behavior and guiding the design of more adaptive and effective reasoning systems. 

One of the foundational concepts in cognitive psychology is metacognition---the ability to monitor and regulate one’s own cognitive processes~\cite{flavell1979metacognition}. In humans, metacognitive skills enable strategic decision-making about when to rely on intuition versus when to engage in more effortful reasoning. 
Complementing this, dual-process theory~\cite{kahneman2011thinking} distinguishes between two modes of thinking: System 1 (fast, automatic, intuitive) and System 2 (slow, effortful, analytical).
These two systems are not mutually exclusive but interact dynamically, with System 2 often regulating or overriding System 1 when deeper reflection is needed.
Cognitive load theory~\cite{sweller1988cognitive} emphasizes that high-complexity tasks that exceed working memory capacity often necessitate the use of external aids or simplified representations to maintain performance.

Analogously, the reasoning process of LLMs benefits from switching among fast, slow, and tool-augmented thinking, balancing efficiency and accuracy depending on task demands. 
However, this analogy should be understood pragmatically rather than as a claim of cognitive equivalence. LLMs do not possess human working memory, phenomenological experience, or biologically grounded automaticity. Therefore, their fast, slow, and tool-augmented thinking should not be interpreted as direct counterparts of human System 1 and System 2 processes. Nevertheless, since LLMs can rely on internal parametric knowledge, engage in deliberate reasoning, and use external resources, cognitive science continues to inspire architectural and algorithmic design of LLMs, shaping how they reason, adapt, and interact in increasingly intelligent ways.

%% file: imex.tex
\begin{figure*}[t]
    \centering
    \includegraphics[width=0.8\textwidth]{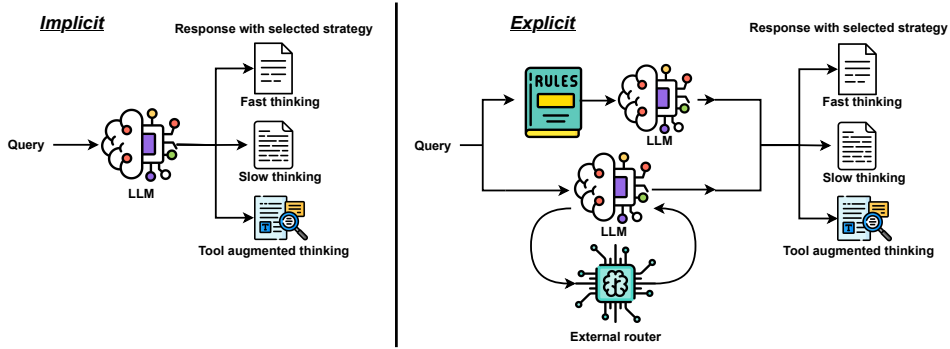}
    \caption{The comparison between implicit and explicit reasoning strategy selection. 
    In implicit selection (left), the reasoning strategy is determined internally by the LLM. In contrast, explicit selection (right) relies on additional rules or external routers to choose the appropriate reasoning strategy.}
    \label{fig:selector}
\end{figure*}

\section{Overall Selection Paradigm}
In this section, we review common paradigms for selecting reasoning strategies in LLMs, as illustrated in Figure~\ref{fig:selector}. 
We group existing approaches into two categories: \textit{implicit} and \textit{explicit} selection.
Implicit selection refers to models that internally determine reasoning strategies, typically learned end-to-end during post-training. In contrast, explicit selection refers to approaches that rely on externally specified mechanisms to guide reasoning strategy choice, such as hand-crafted rules and learned router networks.


From the perspective of our framework in Section~\ref{sec:def}, the two paradigms differ primarily in how selection procedures are instantiated.
Specifically, implicit selection obtains the procedure $\mu$ in our framework (Eq.~\eqref{eq:phi_d} and~\eqref{eq:phi_s}) from the model parameters $\theta$, where this capability is acquired through SFT or RL-based post-training.
On the contrary, explicit selection obtains the procedure from explicit rules or external routers.
Based on the decision scores produced by the procedures, these scores are compared with the thresholds defined by the decision functions in Eq.~\eqref{eq:5} and~\eqref{eq:6}, resulting in the final depth of reasoning and the source of knowledge.

There are several practical differences between the two selection paradigms. ~\textit{Explicit} selection provides fine-grained external control, as predefined rules or external routers determine when to invoke different reasoning strategies. In contrast, ~\textit{implicit} selection is based on internally learned policies, offering smoother integration into the model but less direct controllability. Training cost also differs by design: ~\textit{implicit} selection requires the model to acquire adaptive behavior through SFT or RL during post-training, whereas ~\textit{explicit} selection typically trains a lightweight routing module.

\subsection{Implicit Selection}
For implicit selection, the adaptive selection ability is typically acquired during post-training, where the model learns to associate certain input distributions or task features with distinct reasoning behaviors~\cite{luo2025autol2s,zhang2025adaptthink}.
In particular, two types of post-training methods are adopted: \textit{Supervised Fine-Tuning (SFT)} and \textit{Reinforcement Learning (RL)}.

\subsubsection{Supervised Fine-tuning (SFT)}
Supervised Fine-Tuning (SFT) trains LLMs on curated datasets where distinct reasoning behaviors are embedded within the training samples~\cite{zhao2025let,li2025tl,asai2023self,schick2023toolformer,nakano2021webgpt}. 
Through exposure to diverse task types that span different knowledge boundaries and solution formats, the model implicitly learns to associate input patterns with appropriate reasoning strategies.

For instance, \cite{luo2025autol2s} construct training data that includes both short and long Chain-of-Thought (CoT) reasoning paths, accompanied by a special indicator token signaling the preferred path.
Similar ideas are also explored in Self-Braking Tuning (SBT)~\cite{zhao2025let} and SynapseRoute~\cite{zhang2025synapseroute}.
Additionally, \cite{asai2023self} introduces the use of reflection tokens, enabling the model to dynamically choose whether to invoke tool-augmented reasoning and retrieve the external knowledge.

\subsubsection{Reinforcement Learning (RL)}
Reinforcement Learning (RL) frameworks further enable models to optimize reasoning strategies based on reward signals. 

For example, KAT-V1~\cite{zhan2025kat} introduces a step-SRPO reinforcement learning framework, enabling the model to adaptively switch between reasoning and non-reasoning modes based on the input query. AdaCoT~\cite{lou2025adacot}, trained using the Proximal Policy Optimization (PPO) algorithm, implements adaptive reasoning by determining the necessity of CoT reasoning based on query complexity.
AdaptThink~\cite{zhang2025adaptthink} formulates this as a constrained RL problem, introducing an importance sampling strategy to balance the use of different reasoning strategies.
LHRMs~\cite{jiang2025think} further explore hybrid reasoning by combining hybrid fine-tuning with online RL using Hybrid Group Policy Optimization (HGPO).
AutoThink~\cite{tu2025learning} adopts a multi-stage RL framework with reward shaping to learn policies that jointly optimize reasoning quality and computational cost. 

\vspace{0.5em}
\subsection{Explicit Selection}
Explicit selection introduces external mechanisms that explicitly guide the model in choosing an appropriate reasoning strategy. 
We further categorize existing methods into \textit{rule-based} and \textit{model-based} selecting approaches.

\subsubsection{Rule-based Selection}
Rule-based selection refers to approaches that rely on manually defined heuristics or fixed criteria to determine the appropriate reasoning strategy, typically according to model confidence, task complexity, or input characteristics.

For example, \cite{kojima2022large} point out that a simple prompt like ``Let's think step by step'' can activate CoT reasoning, improving performance on complex tasks.
Later, \cite{yan2025mur} propose a test-time control mechanism that uses a momentum-based uncertainty threshold to determine whether to engage in deeper reasoning strategies.
Rule-based Confidence Reasoning (RCR)~\cite{huang2024trust} explicitly compares the model's internal-only and external context-based answers using a confidence scoring function, and selects the answer with higher confidence as the final output.

The rule-based methods are typically lightweight, interpretable, and easy to integrate into existing inference pipelines. However, their effectiveness relies heavily on the quality and generality of the handcrafted rules.

\subsubsection{Model-based Selection}

Model-based selection refers to approaches that employ a learned external router to automatically determine the most appropriate reasoning strategy based on the input query~\cite{research2025exaone,he2025self,liang2025thinkswitcher}. 

ThinkSwitcher~\cite{liang2025thinkswitcher} enables a single LLM to adaptively switch between short chain-of-thought (SC) and long chain-of-thought (LC) reasoning modes. It trains a lightweight switching module that, given a query embedding, predicts the empirical pass rates of both modes and selects LC if its predicted success rate exceeds that of SC by a predefined threshold.
Similarly, Self-Route~\cite{he2025self} introduces a lightweight classifier trained on the model’s internal hidden representations to serve as a router. This classifier dynamically selects between reasoning modes during inference, allowing the model to adjust its strategy based on input complexity or uncertainty.

Compared to the rule-based methods, learned external routers can capture complex, data-driven knowledge boundaries that may be difficult to encode manually. 

\subsection{Discussion}
Adaptive reasoning strategy selection generally aims at allocating reasoning effort which yields the highest marginal value.

Implicit selection enables LLMs to autonomously infer and adopt appropriate reasoning strategies without relying on external controllers or routing mechanisms.
A central challenge for implicit selection methods lies in: how to effectively construct and compose the training data across different reasoning strategies.
Improper data composition can bias the LLMs toward overusing certain strategies or failing to generalize across diverse tasks.


Explicit selection often adds an external router that turns this problem into a system-level policy over strategies, tools, and even models.
In practice, many production systems employ learned routers to decide when to invoke heavier reasoning or specialized components. For example, GPT-5 series~\cite{openai2025gpt5} represents a production-scale explicit selection framework. 
Its interface includes Instant, Thinking, and Auto modes, allowing users to manually control reasoning depth. Instant is typically used for routine or lightweight tasks, while Thinking is intended for more complex problems, corresponding to fast and slow thinking, respectively. In addition, an Auto mode employs a real-time router that selects among reasoning modes based on conversation type, complexity, tool requirements, and user intent.
However, explicit selection can introduce orchestration overhead, and the router’s estimates may be misaligned with base-model capabilities---both underestimation and overestimation can add errors or costs.

%% file: FCS-251673-fig3.tex

\definecolor{paired-light-blue}{RGB}{193, 229, 245}
\definecolor{paired-dark-blue}{RGB}{17, 121, 162}

\definecolor{paired-light-orange}{RGB}{251, 208, 162}
\definecolor{paired-dark-orange}{RGB}{230, 85, 12}
\definecolor{paired-light-green}{RGB}{215, 250, 245}
\definecolor{paired-dark-green}{RGB}{114, 139, 134}
\definecolor{paired-light-purple}{RGB}{218, 218, 235}
\definecolor{paired-dark-purple}{RGB}{117, 107, 176}
\definecolor{paired-light-gray}{RGB}{217, 217, 217}
\definecolor{paired-dark-gray}{RGB}{99, 99, 99}
\definecolor{paired-light-pink}{RGB}{222, 158, 214}
\definecolor{paired-dark-pink}{RGB}{123, 65, 115}
\definecolor{paired-light-red}{RGB}{231, 150, 156}
\definecolor{paired-dark-red}{RGB}{131, 60, 56}
\definecolor{paired-light-yellow}{RGB}{231, 204, 149}
\definecolor{paired-dark-yellow}{RGB}{141, 109, 49}
\definecolor{light-green}{RGB}{118, 207, 180}
\definecolor{raspberry}{RGB}{228, 24, 99}

\tikzset{%
    root/.style =          {align=center,text width=3cm,rounded corners=3pt, line width=0.5mm, fill=paired-light-gray!50,draw=paired-dark-gray!90},
    env_section/.style =  {align=center,text width=4cm,rounded corners=3pt, fill=paired-light-green!80,draw=paired-dark-green!100,line width=0.4mm},
    capability_section/.style = {align=center,text width=4cm,rounded corners=3pt, fill=paired-light-purple!80,draw=paired-dark-purple!100,line width=0.4mm},
    training_section/.style = {align=center,text width=4cm,rounded corners=3pt, fill=paired-light-green!50,draw=paired-dark-green!80, line width=0.4mm},
    inference_section/.style = {align=center,text width=4cm,rounded corners=3pt, fill=paired-light-red!35,draw=paired-light-red!90, line width=0.4mm},
    protocol_section/.style = {align=center,text width=4cm,rounded corners=3pt, fill=paired-light-yellow!40,draw=paired-dark-yellow!100, line width=0.4mm},
    discussion_section/.style = {align=center,text width=4cm,rounded corners=3pt, fill=paired-light-red!20,draw=paired-dark-red!100, line width=0.4mm},
    general_section/.style = {align=center,text width=4cm,rounded corners=3pt, fill=paired-light-purple!35,draw=paired-dark-purple!90, line width=0.4mm},
    subsection/.style =    {align=center,text width=3.5cm,rounded corners=3pt}, 
}

\begin{figure*}[!htb]
    \centering
    \resizebox{1\textwidth}{!}{
    \begin{forest}
        for tree={
            forked edges,
            grow'=0,
            draw,
            rounded corners,
            node options={align=center, anchor=center},
            text width=4cm,
            s sep=6pt,
            calign = center,
            calign child=(n_children()+1)/2,
            l sep=12pt,
        },
        [Reasoning Strategy Selection, root,
            [Fast/Slow Knowledge Boundary (\S\ref{sec:fast_slow}), env_section,
                [Model Confidence \\(\S\ref{sec:confidence_fs}), env_section 
                [
                    DynaThink~\cite{pan2024dynathink}; 
                    UnCert-CoT~\cite{zhu2025uncertainty}; 
                    SBT~\cite{zhao2025let}; 
                    MUR~\cite{yan2025mur}; 
                    Early-Exit~\cite{zhang2025reasoning}; 
                    DEER~\cite{yang2025dynamic}; 
                    ThinkNoThink~\cite{wu2025thinking} 
                    ,env_section, text width=12cm
                    ] 
                ]
                [Task Complexity \\(\S\ref{sec:complexity_fs}), env_section
                    [
                    Z1~\cite{yu2025z1}; 
                    AdaCoT~\cite{lou2025adacot};
                    AdaptThink~\cite{zhang2025adaptthink}; 
                    ThinkSwitcher~\cite{liang2025thinkswitcher}; 
                    HGPO~\cite{jiang2025think};
                    SelfRoute~\cite{he2025self}; 
                    AutoL2S~\cite{luo2025autol2s}; 
                    TLDR~\cite{li2025tl}; 
                    SynapseRoute~\cite{zhang2025synapseroute}; 
                    KAT-V1~\cite{zhan2025kat} 
                    Qwen3~\cite{yang2025qwen3};
                    Llama-Nemotron~\cite{bercovich2025llama}; 
                    ,env_section, text width=12cm
                    ] 
                ]
            ]
            [Internal/External Knowledge Boundary (\S\ref{sec:internal_external}), capability_section,
                [Model Confidence \\ (\S\ref{sec:confidence_ie}), capability_section 
                    [
                    FLARE~\cite{jiang-etal-2023-active}; 
                    UniWeb~\cite{li2023web}; 
                    SelfRag~\cite{asai2023self}; 
                    DRAGIN~\cite{su2024dragin}; 
                    SUGAR~\cite{zubkova2025sugar}; 
                    Seakr~\cite{yao2024seakr}; 
                    Rowen~\cite{ding2024retrieve}; 
                    MetaTrigger~\cite{li2025adaptive}; 
                    Alignment~\cite{xu2025alignment} 
                    ,capability_section, text width=12cm] 
                ]
                [Utility Gain\\ (\S\ref{sec:utility_ie}), capability_section 
                    [
                    Toolformer~\cite{schick2023toolformer}; 
                    AWL~\cite{lyu2024adapting}; 
                    ToCodeEM~\cite{wang2025code}; 
                    ReTool~\cite{feng2025retool}; 
                    R3-RAG~\cite{li2025r3}; 
                    ARITIST~\cite{singh2025agentic};
                    RAGate~\cite{wang2024adaptive} 
                    Toolken+~\cite{yakovlev2024toolken+} 
                    ,capability_section, text width=12cm] 
                ]
            ]
        ]
    \end{forest}
    }
    \caption{Decision factors for adaptive reasoning strategy selection. } 
    \label{fig:taxonomy}
\end{figure*}
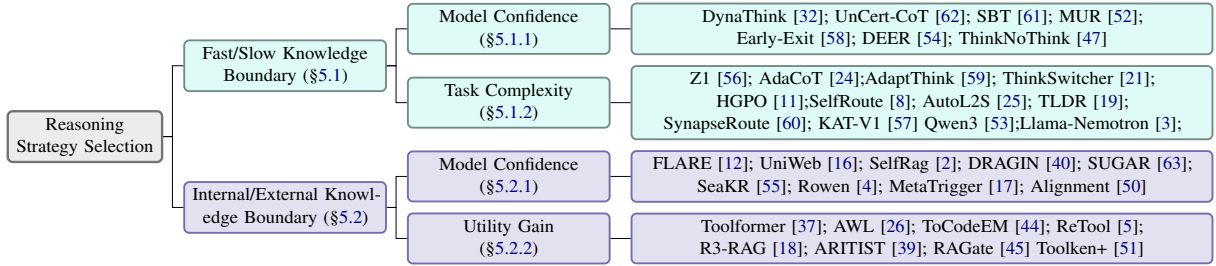

%% file: knowledgeboundary.tex
\section{Decision Factors in Strategy Selection}

In this section, we present a detailed review of existing literature through the lens of the key decision factors governing the choice of reasoning strategies. To this end, we develop a systematic taxonomy organized around two knowledge boundaries, as illustrated in Figure 3.

The intersection of computation depth and knowledge source yields four distinct methodological categories: Fast-Internal, Slow-Internal, Fast-External, and Slow-External. The Fast-Internal category corresponds to standard direct inference without explicit deliberation. Because reasoning is fundamentally designed to surpass the limitations of direct inference, the majority of current research concentrates on Slow-Internal and Slow-External approaches. Within these paradigms, models allocate additional computational resources or consult external tools to resolve complex queries. By contrast, the Fast-External category, characterized by immediate access to external resources without deliberate reasoning, has been less systematically studied as a dedicated research direction. Nevertheless, it is commonly reflected in production style systems such as function calling frameworks, where an LLM invokes predefined functions or APIs and incorporates the returned results, and simple RAG pipelines, where top ranked external passages are retrieved and directly provided as context without explicit reasoning chains. Acknowledging these cases makes the four quadrant taxonomy more comprehensive.

\subsection{Fast/Slow Knowledge Boundary}
\label{sec:fast_slow}
The fast/slow knowledge boundary determines when a model should rely on direct responses versus engaging in longer, more computationally intensive reasoning. 
Two primary decision factors at this knowledge boundary are \textit{model confidence} and \textit{task complexity}. 
In general, high confidence and low complexity are processed by the fast reasoning strategy, whereas uncertainty or greater complexity are handled by the slow reasoning strategy.

\subsubsection{Model Confidence}
\label{sec:confidence_fs}
Model confidence captures the LLM’s estimated likelihood that the current output is correct.
Confidence guided selection dynamically adjusts reasoning strategies based on the model's internal confidence estimates, thereby optimizing the trade-off between performance and computational cost.

CRL~\cite{pawitan2025confidence} provides one of the first in-depth analyses of LLMs' self-reported and behavioral confidence on reasoning tasks, showing a positive but imperfect correlation between confidence and accuracy. 

Building on this, recent work has started to explore confidence-guided decision frameworks that dynamically switch between fast and slow reasoning. 
DynaThink~\cite{pan2024dynathink} proposes two selection criteria to guide the choice between fast and slow reasoning: consistency verification and reasoning depth. In our Section 2 formulation, these criteria can be interpreted as the decision features used in the selection function. The framework selects fast thinking when a single answer achieves high consistency and low reasoning depth, which in our formulation corresponds to the decision score crossing the selection threshold. Problems that fail either condition default to slow thinking, involving more resource‐intensive and deliberate analysis.




For code generation, UnCert-CoT~\cite{zhu2025uncertainty} proposes entropy-based and probability-differential-based measures of confidence. When confidence is low, UnCert-CoT triggers long CoT decoding to sample multiple reasoning paths and selects the final code with the highest estimated likelihood of correctness. 
SBT~\cite{zhao2025let} adopts a reasoning efficiency ratio and an overthinking label ratio to detect overthinking, enabling early termination when thresholds are met. 
Similarly, MUR~\cite{yan2025mur} aggregates historical step-level uncertainties to assess the confidence of a reasoning path and selectively allocates computation to the most critical steps.

Another line of work focuses on self-truncation based on model confidence during reasoning.
Early-Exit~\cite{zhang2025reasoning} finds models’ hidden states encode the correctness of future answers, enabling early prediction of the correctness and deciding whether to exit reasoning.
DEER~\cite{yang2025dynamic} monitors transitions in reasoning, and dynamically exits when the confidence of the trial answer is sufficiently high.
More recently, ThinkNoThink~\cite{wu2025thinking} jointly generates fast and slow reasoning steps and stops reasoning when two responses are consistent.

Collectively, these studies demonstrate that confidence estimation---whether derived from token probabilities, entropy, or the self-consistency~\cite{wang2022self} of reasoning paths---offers a powerful control signal for adaptive reasoning. Nonetheless, while higher confidence often correlates with better performance, LLMs can be overconfident, which can misguide strategy selection~\cite{pawitan2025confidence}.

\subsubsection{Task Complexity}
\label{sec:complexity_fs}
Task complexity captures the intrinsic difficulty of an input query, typically estimated by how much reasoning depth, exploration, or verification is likely required to reach a right answer. 
Unlike confidence, which is an internal belief about an answer, task complexity is an ex-ante property of the problem.

To build supervision for complexity-aware selection, Z1~\cite{yu2025z1} curates a dataset that integrates both short reasoning trajectories for simple problems and strong reasoning trajectories for complex problems, where complexity is determined by the Evol-Instruct method~\cite{xu2024wizardlm} to span the difficulty spectrum.
Methods like AutoL2S~\cite{luo2025autol2s}, SynapseRoute~\cite{zhang2025synapseroute}, and KAT-V1~\cite{zhan2025kat} seek to explore automated annotation methods to label queries as ``thinking'' or ``non-thinking'' by applying a rule-based evaluation of answer accuracy, inference time, and token usage.
TLDR~\cite{li2025tl} further introduces a dynamic re-weighting method that continuously adjusts the training data ratio between fast and slow thinking.

Recent work uses RL to endow models with internal control over reasoning depth. 
AdaCoT~\cite{lou2025adacot} formulates adaptive reasoning as a Pareto optimization problem, seeking to balance model performance with the computational costs.
It employs an RL policy to dynamically assess the complexity of incoming queries and determine the necessity and extent of the reasoning steps.
In the meanwhile, AdaptThink~\cite{zhang2025adaptthink} employs a constrained optimization objective within an on-policy RL framework, teaching the model to adaptively switch to a fast thinking mode for simpler problems to reduce inference costs without sacrificing accuracy.
The Hybrid Group Policy Optimization (HGPO) algorithm~\cite{jiang2025think} is an online RL method designed to enable the model to implicitly learn when to select the appropriate reasoning strategy. HGPO is guided by comparing the relative quality of fast and slow thinking responses for a given query, using inter-group and intra-group reward signals.

A complementary approach externalizes the decision via lightweight routers.
ThinkSwitcher~\cite{liang2025thinkswitcher} deploys a lightweight external router that dynamically selects between fast and slow thinking modes based on task complexity.
The router is trained with supervision signals derived from the empirical pass rates of the backbone model executing both reasoning modes.
Similarly, Self-Route~\cite{he2025self} also trains a lightweight external router to estimate the model's own capability for different levels of task complexity by analyzing its hidden layer representations. 
The router then dynamically decides whether to use slow thinking for the current problem at inference time.
The levels of task complexity are computed by the accuracy of a proxy LLM.

While some recent LLMs, such as Qwen3~\cite{yang2025qwen3} and Llama-Nemotron\cite{bercovich2025llama}, support hybrid reasoning modes, they still rely on user-specified reasoning strategies rather than enabling automatic reasoning-mode selection.
Overall, complexity-based selection offers a robust and scalable framework for adaptive reasoning by explicitly tailoring model behavior to input difficulty. 

\subsection{Internal/External Knowledge Boundary}
\label{sec:internal_external}

The internal/external knowledge boundary concerns the extent to which a model can rely on its internal parametric knowledge to solve a given task, versus when it must retrieve external knowledge sources with tool-augmented thinking. Key decision factors at this boundary are the \textit{model confidence} and the \textit{utility gain} from using an external tool.

\subsubsection{Model Confidence}
\label{sec:confidence_ie}
Using model confidence to navigate between the internal/external knowledge boundary leverages an LLM's own uncertainty to select the appropriate knowledge source. The core intuition is that an LLM can monitor its certainty about its own outputs: high confidence suggests the model can answer correctly using its internal knowledge, whereas low confidence triggers a switch to retrieval, computation, or another external resource. 

One direction focuses on uncertainty modeling via output probabilities. The confidence is derived directly from the model’s output probabilities, such as token-level likelihoods~\cite{jiang-etal-2023-active, su2024dragin,asai2023self,zubkova2025sugar}, or output entropy~\cite{li2023web}, then compared with a threshold to decide on tool usage. 
Notably, SUGAR~\cite{zubkova2025sugar} addresses the challenge that in natural language, the same idea can be expressed in various syntactic and lexical forms. It proposes using semantic entropy to determine the knowledge boundary, measuring uncertainty over meaning rather than specific token sequences. In our Section 2 formulation, semantic entropy serves as a confidence-related decision feature in the selection function, where higher semantic entropy indicates lower confidence and triggers RAG.
Although these methods are simple and efficient, their performance depends heavily on calibration: overconfident models may skip necessary retrieval, while underconfident models may trigger excessive tool use \cite{schuster2022confident, mialon2023gaiabenchmarkgeneralai}.

A recent direction is to exploit consistency across multiple generated samples or within the model's latent space~\cite{xu2025alignment}. 
Instead of relying solely on output probabilities, these methods examine the model’s hidden representations or the variability across different reasoning paths to better gauge its true certainty.
Seakr~\cite{yao2024seakr}, for instance, measures the consistency of an LLM's internal hidden states across multiple generations for the same prompt to decide when to retrieve external knowledge.
Rowen~\cite{ding2024retrieve} takes a similar approach by generating semantically equivalent perturbations of a query and measuring the consistency of answers across different languages and models. A low consistency score, indicating a potential hallucination, triggers the retrieval of external knowledge for correction.
MeCo~\cite{li2025adaptive} instead derives a meta-cognitive score of consistency with a lightweight representation probe and applies a dual-threshold policy to decide whether to invoke external tools.
While these methods can better capture a model’s knowledge gaps, they often require access to model internals or multiple inference passes.

\subsubsection{Utility Gain}
\label{sec:utility_ie}
Selection methods that use utility gain as a decision factor decide whether to retrieve external information or invoke tools by estimating the expected benefit of doing so---\eg, reductions in perplexity, or improvements in accuracy.

Toolformer~\cite{schick2023toolformer} formulates external tool use as utility-driven modeling by executing candidate API calls and retaining only those that reduce the perplexity of subsequent tokens. This self-supervised filtering implicitly learns a policy that adaptively triggers external tools without sacrificing the general abilities of LLMs. 
For scientific problem solving, AWL~\cite{lyu2024adapting} leverages the solvability of a sample without tools as the decision factor, guiding the model to answer easy queries internally while invoking tools on hard ones.

More recently, RL-based approaches have been used to endow LLMs with adaptive tool selection.
For mathematical problem-solving, ToCodeEM~\cite{wang2025code} employs an Expectation-Maximization (EM) framework where the E-step performs guided exploration, estimating the expected success of code-integrated versus pure reasoning paths through rollouts. The M-step then uses these utility estimates in an off-policy RL objective to jointly optimize tool usage and reasoning.

In addition, ReTool~\cite{feng2025retool} treats tool invocation as an action within a continuous reasoning process and optimizes the policy with a simple final-correctness reward. The model autonomously discovers how to interleave textual reasoning with code execution to maximize expected task accuracy.
R3-RAG~\cite{li2025r3} combines a fine-grained process reward for the relevance of each retrieval step, and a final outcome reward for the correctness of the answer with an RL framework. This enables the model to learn an adaptive policy for interleaving its internal reasoning with external knowledge retrieval.
Meanwhile, ARTIST~\cite{singh2025agentic} introduces a unified framework where models learn to dynamically interleave internal reasoning with external tool use as a coherent thinking process. optimized by GRPO.

Another line of work uses lightweight external routers to endow models with adaptive tool-use capabilities. 
For example, RAGate~\cite{wang2024adaptive} employs a lightweight router that adaptively selects only internal thinking or tool-augmented thinking, trained with supervision labels derived from human preference.
Toolken+~\cite{yakovlev2024toolken+} further improves adaptive tool selection by learning a lightweight reranker that considers the top-k tool candidates alongside a special ``Reject'' option.

\begin{table*}[htbp] 
\centering
\caption{Summary of Representative Reasoning Strategy Selection Methods}
\small
\setlength{\tabcolsep}{2pt}
\begin{tabular}{@{}%
    >{\RaggedRight\arraybackslash}m{0.16\textwidth} 
    >{\RaggedRight\arraybackslash}m{0.14\textwidth} 
    >{\centering\arraybackslash}m{0.10\textwidth} 
    >{\RaggedRight\arraybackslash}m{0.28\textwidth} 
    >{\centering\arraybackslash}m{0.06\textwidth}   
    >{\RaggedRight\arraybackslash}m{0.16\textwidth} 
    >{\RaggedRight\arraybackslash}m{0.10\textwidth} 
  @{}}
\toprule
\textbf{Work} & \textbf{Boundary Type} & \textbf{Decision Factor} & \textbf{Base Model} & \textbf{Open Source} & \textbf{Application} & \textbf{Selection Paradigm} \\
\midrule
DynaThink~\cite{pan2024dynathink} &
Fast/Slow &
Model Confidence &
GPT-3.5 Turbo; GPT-4; Gemini; Mixtral-8x7B &
Yes &
Math reasoning; Commonsense QA &
Explicit \\
\hline

UnCert-CoT~\cite{zhu2025uncertainty} &
Fast/Slow &
Model Confidence &
Qwen2.5-Coder-7B-Base; deepSeek-coder-6.7B-base &
No &
Code generation &
Explicit \\
\hline

SBT~\cite{zhao2025let} &
Fast/Slow &
Model Confidence &
Qwen2.5-Math-1.5B/7B-Instruct; Llama-3.2-1B-Instruct; Llama-3.1-8B-Instruct &
Yes &
Math reasoning &
Explicit \\
\hline

MUR~\cite{yan2025mur} &
Fast/Slow &
Model Confidence &
Qwen3-1.7B/4B/8B &
Yes &
Math reasoning; Science QA &
Explicit \\
\hline

Early-Exit~\cite{zhang2025reasoning} &
Fast/Slow &
Model Confidence & 
R1-Distill-Llama-8B/70B; R1-Distill-Qwen-1.5B/7B/32B; QwQ-32B & 
Yes & 
Math reasoning; Logic QA &
Explicit \\ 
\hline

Z1~\cite{yu2025z1}&
Fast/Slow &
Task Complexity &
Qwen2.5-Coder-7B-Instruct  &
Yes &
Code reasoning; Math; Science QA &
Implicit \\
\hline

AdaCoT~\cite{lou2025adacot} &
Fast/Slow &
Task Complexity &
Internal MoE 15B/150B (ByteDance) &
No &
Math reasoning; General QA; Exams &
Implicit \\
\hline

ThinkSwitcher~\cite{liang2025thinkswitcher} &
Fast/Slow &
Task Complexity &
DeepSeek-R1-Distill-Qwen-1.5B/7B/14B &
No &
Math reasoning &
Explicit \\
\hline

TLDR~\cite{li2025tl} &
Fast/Slow &
Task Complexity &
DeepSeek-R1-Distill-7B/14B &
No &
Math reasoning &
Implicit\\
\hline

SynapseRoute~\cite{zhang2025synapseroute} &
Fast/Slow &
Task Complexity &
Qwen3-30B-a3b &
No &
Medical QA &
Implicit \\
\hline

FLARE~\cite{jiang-etal-2023-active} &
Internal/External &
Model Confidence  &
text-davinci-003 &
Yes &
Multihop QA; Commonsense reasoning &
Explicit \\

\hline

SelfRag~\cite{asai2023self} & Internal/External & Model Confidence &
Llama-2-7B/13B & Yes & Open-domain QA; Fact verification & Implicit \\
\hline

SUGAR~\cite{zubkova2025sugar} &
Internal/External &
Model Confidence &
Llama-2-chat-7B; contriever-msmarco &
No &
Open-domain QA &
Explicit \\
\hline

Seakr~\cite{yao2024seakr} &
Internal/External &
Model Confidence &
Llama-2-7B-Base/Chat; Llama-3-8B-Base/Instruct &
Yes &
Multi-hop QA; Open-domain QA &
Explicit \\
\hline

Alignment~\cite{xu2025alignment} &
Internal/External &
Model Confidence &
Llama-3.1-8B-Instruct; Qwen-2.5-7B-Instruct; DeepSeek-R1-671B  &
No &
Arithmetic; Open-domain QA &
Explicit \\
\hline

Toolformer~\cite{schick2023toolformer} &
Internal/External &
Utility Gain &
GPT-J 6.7B  &
No &
Open-domain QA; Math reasoning &
Implicit \\
\hline

AWL~\cite{lyu2024adapting} &
Internal/External &
Utility Gain &
Llama-3.1-8B-Instruct / Qwen2.5-14B-Instruct &
Yes &
Science QA; Math Solving &
Implicit \\
\hline

ToCodeEM ~\cite{wang2025code} &
Internal/External &
Utility Gain &
Qwen2-Math-7B; Qwen-2.5-Base-7B; DeepSeekMath-Instruct-7B &
Yes &
Math reasoning &
Implicit \\
\hline

ReTool~\cite{feng2025retool} &
Internal/External &
Utility Gain &
Qwen2.5-32B-Instruct / DeepSeek-R1-Distill-Qwen-32B &
Yes &
Math reasoning &
Implicit \\
\hline

RAGate~\cite{wang2024adaptive} &
Internal/External &
Utility Gain &
Llama-2-7B / 13B / 70B; GPT-2; BERT-ranker; TF-IDF &
Yes &
Conversational systems &
Explicit \\

\bottomrule
\end{tabular}
\label{tab:survey_summary4}
\end{table*}




\vspace{0.5em}
\noindent{\normalfont\normalsize 5.3\hspace{1em}Discussion}\par
\vspace{0.5em}

Different decision factors capture distinct but complementary aspects of the reasoning strategy selection problem. These factors can be integrated to enable more robust and adaptive decision-making. 
Specifically, confidence reflects the model's certainty in its current response; complexity estimates the inherent difficulty of the input problem; and utility gain evaluates the expected payoff of allocating additional computational resources. 
However, relying on these decision factors introduces specific failure modes if they are inaccurately estimated. Most notably, when confidence estimates are poorly calibrated, overconfident models tend to skip necessary external retrieval or internal deliberation, leading to premature and potentially hallucinated responses~\cite{ni2024llms, huang2025survey}. On the other hand, underconfident models may invoke excessive reasoning steps or redundant tool use, introducing severe latency and unnecessary computational overhead. Furthermore, poorly estimated complexity or utility gain can result in a fundamental mismatch between the chosen reasoning strategy and the problem's actual requirements, compromising both accuracy and efficiency.

Notably, these decision factors closely align with human metacognitive processes, which involve monitoring and regulating one’s own cognitive activity~\cite{kahneman2011thinking}. Just as humans adjust their reasoning strategies based on perceived confidence, task difficulty, and the potential payoff of further effort, language models can benefit from the similar mechanisms to dynamically select between fast, slow, or tool-augmented reasoning.  Table~\ref{tab:survey_summary4} provides a summary of existing methods and their application across different knowledge boundaries.

%% file: future.tex
\section{Summary and Future Prospects}
Adaptively selecting LLM reasoning strategies has continued to be an inspiring domain.
Recent studies have begun to explore mechanisms that enable dynamic alignment of reasoning modes with the demands of specific tasks. As LLM capabilities continue to scale, their applications are rapidly expanding across critical domains such as education, healthcare, law, science, and engineering~\cite{liu2025real}. Developing efficient reasoning strategies can greatly improve their usability and trustworthiness, helping to reduce unnecessary computational costs and minimize hallucinations.

Looking ahead, we envision several promising directions for future research:
\paragraph{Pre-training/Mid-training Foundation:} 
While most existing work focuses on post-training mechanisms for reasoning strategy selection, it is during pre-training that LLMs acquire most of their internal knowledge and foundational cognitive capacities. 
This means that adaptive reasoning capabilities are primarily instilled during post-training, which can lead to superficial boundary awareness or catastrophic forgetting of general knowledge. To address this, future work should explore how knowledge boundaries can be explicitly modeled much earlier, during pre-training or mid-training. The field must determine whether this boundary awareness can be deeply integrated into large-scale pre-training without harming the model's general capabilities or downstream adaptability.

\paragraph{Unified Selection:} Existing strategy selection methods often lack the capability to integrate multiple reasoning strategies in a seamless, human-like manner. 
More specifically, current approaches typically treat internal reasoning depth and external tool invocation as isolated, disjointed decisions, preventing models from fully optimizing their computational resources. A promising direction is to develop models that autonomously assess task complexity and dynamically select the optimal combination of strategies at each step. How can we jointly model these orthogonal dimensions to enable seamless, unified transitions across reasoning strategies?

\paragraph{Reasoning Orchestration Systems:} 
As multi-agent frameworks become more prevalent~\cite{tran2025multi}, current orchestration systems often struggle to resolve inter-agent conflicts and suffer from compounding inefficiencies as the system scales. Future frameworks must determine how to effectively delegate subtasks and ensure coherence in collective reasoning. A major challenge moving forward will be designing orchestration policies that maintain efficiency, interpretability, and robustness even as the number of agents and the complexity of their interactions increase.

\paragraph{Multimodal Selection:} Human reasoning often integrates multiple modalities such as language, vision, action, and spatial awareness~\cite{lehmann2020interaction}. 
However, current adaptive reasoning research focuses heavily on text-based tasks, and models lack the flexibility to dynamically shift between modalities when solving complex, real-world problems. Future research must investigate how to effectively fuse and select among diverse reasoning modalities, including text, images, audio, video, and physical execution. Researchers have yet to answer how a model can dynamically evaluate and coordinate the most appropriate reasoning strategies across these different modalities when confronted with interleaved multimodal tasks.

\paragraph{Personalized Reasoning:} Different users may require different styles or levels of reasoning depending on their goals, expertise, or preferences. 
In practice, most existing LLMs apply a one-size-fits-all reasoning approach, failing to adjust the depth, formality, or transparency of their reasoning chains based on specific user profiles or contexts. Developing models capable of personalizing their reasoning behavior is a highly promising avenue. However, it remains an open question how to reliably evaluate and support this personalization in highly specialized domains—such as sports coaching or legal advice—where task-specific training data and gold-standard evaluation criteria are severely limited.

\paragraph{Robustness and Safety:} As reasoning systems become more adaptive and autonomous, ensuring their robustness and safety becomes increasingly critical. 
With greater autonomy over their reasoning strategies, models often become vulnerable to blindly trusting unreliable external tools or failing to recognize when their adaptive strategies are compromised by distribution shifts. Ensuring robustness involves handling ambiguous inputs, avoiding unethical conclusions, and maintaining human alignment. What mechanisms can be developed to allow adaptive reasoning systems to autonomously detect when strategy switching or external tool invocation becomes unsafe?